\documentclass[]{spie}  

\usepackage{amsmath,amsfonts,amssymb}
\usepackage{graphicx}
\usepackage{subcaption}
\usepackage{bm}
\usepackage{tikz}
\usepackage[colorlinks=true, allcolors=blue]{hyperref}
\usepackage{algorithm}
\usepackage{algpseudocode}

\newcommand{\R}{\mathbb{R}}

\newcommand{\C}{\mathbb{C}}

\newcommand{\br}{\boldsymbol{\rho}}

\newcommand{\FF}{\mathbf{F}}

\newcommand{\GG}{\mathbf{G}}
\newcommand{\RR}{\mathbf{R}}
\newcommand{\HH}{\mathbf{H}}
\newcommand{\dd}{\mathbf{d}}
\newcommand{\cc}{\mathbf{c}}
\newcommand{\bb}{\mathbf{b}}

\newcommand{\xx}{\mathbf{x}}
\newcommand{\bx}{\boldsymbol{x}}

\newcommand{\AAA}{\mathbf{A}}

\title{Online CS-based SAR Edge-Mapping}

\author[a]{Conor Flynn}
\author[a]{Radoslav Ivanov}
\author[a]{Birsen Yaz\i c\i}
\affil[a]{Rensselaer Polytechnic Institute, Troy, New York, USA}

\authorinfo{Further author information: (Send correspondence to Conor Flynn)\\Conor Flynn: E-mail: flynnc3@rpi.edu}

\begin{document} 
\maketitle

\begin{abstract}
With modern defense applications increasingly relying on inexpensive, small Unmanned Aerial Vehicles (UAVs), a major challenge lies in designing intelligent and computationally efficient onboard Automatic Target Recognition (ATR) algorithms to carry out operational objectives. This is especially critical in Synthetic Aperture Radar (SAR), where processing techniques such as ATR are often carried out post data collection, requiring onboard systems to bear the memory burden of storing the back-scattered signals. To alleviate this high cost, we propose an online, direct, edge-mapping technique which bypasses the image reconstruction step to classify scenes and targets. Furthermore, by reconstructing the scene as an edge-map we inherently promote sparsity, requiring fewer measurements and computational power than classic SAR reconstruction algorithms such as backprojection.
\end{abstract}


\section{INTRODUCTION}
As modern defense applications shift toward relying on inexpensive, autonomous Unmanned Aerial Vehicles (UAVs), the challenge of designing computationally efficient and reliable onboard systems becomes increasingly critical \cite{coffey2002emergence, lehtotextordmasculine2021small}. In Synthetic Aperture Radar (SAR), this objective traditionally involves storing and processing received signal data to produce a high-quality image of a scene's reflectivity, which is then used for downstream tasks such as Automatic Target Recognition (ATR).\cite{el2016automatic} However, traditional ATR techniques are typically performed on the reconstructed image rather than directly on the back-scattered signal. This reconstruction adds an additional layer of processing and memory overhead. We propose an alternative to this standard framework by performing image interpretation directly on raw signal data through edge-mapping and compressive sensing, promoting an efficient framework for real-time ATR.

Direct image interpretation involves processing back-scattered signal data to extract features without a full image reconstruction phase. By extracting edges from the raw signal, we eliminate the computational and memory overhead required to generate human-interpretable imagery. While some literature explores direct interpretive approaches\cite{yanik2011computationally, kazemi2019deep}, most research still applies ATR methods to traditional SAR imagery \cite{el2016automatic, deng2017sar, flynn2025comparability, pei2017sar}. Existing direct approaches differ from our proposed method by either focusing solely on edge segmentation\cite{yanik2011computationally} or using deep learning frameworks for raw signal classification\cite{kazemi2019deep}, whereas our framework combines this direct signal interpretation with compressive sensing (CS) for increased computational efficiency and interpretability. CS techniques address ill-posed inverse problems and recover sparse signals from a limited number of measurements. Sparse coding is particularly effective in  sparse SAR, where the goal is to reconstruct information from a limited number of pulses by exploiting signal sparsity.\cite{xu2022sparse, souganlui2014dictionary, flynn2026online}.

Our direct, online, CS-based edge-mapping approach offers several key benefits over traditional sparse SAR processing techniques. First, by leveraging an online compressive sensing algorithm, we significantly reduce the onboard memory required for pulse data storage. Computational costs are further reduced by the direct signal interpretation, removing the need for traditional image reconstruction. This also makes online ATR possible, allowing for target recognition to occur mid-mission rather than requiring downstream tasking post-collection. Also, by combining edge-mapping with CS, we promote a more effective sparse representation that reduces the pulse requirement while retaining high quality information required by ATR. Finally, by transitioning the reconstruction to an online framework, we allow the system to adaptively tune imaging parameters (such as RF-band, look direction, and pulse intervals) to further optimize data retrieval.

The outline of the remainder of the paper is as follows: Section~\ref{sec:radar-imaging} introduces the radar imaging problem in the context of mono-static SAR. Section~\ref{sec:edge-mapping} proposes a direct edge-mapping filter to be applied to the received signal data. Section~\ref{sec:sparse-sar} introduces our online coding schema in the context of sparse SAR.  Finally, Section~\ref{sec:conclusion} summaries our proposal and discusses the future direction of this work.


\section{RADAR IMAGING}
\label{sec:radar-imaging}
To simplify the problem, we assume the start-stop approximation and the far-field assumption hold \cite{carrara1995soptlight}.

Let $\Psi:\R^2\rightarrow \R$ be the ground topography. A scatterer's position in 3D-continuous space is denoted by $\xx\in\R^3$, mapped from its 2D-coordinates $\bx\in\R^2$ such that $\xx=[\bx,\Psi(\bx)]$. The antenna follows a trajectory $\gamma:\left\{s_0,...,s_n,...,s_L\right\}\rightarrow\R^3$, recording its position at each slow-time instance $s_n$. Our objective in this SAR imaging framework is to reconstruct the scene's reflectivity function $\rho:\R^2\rightarrow\R$, achieved by analyzing the back-scattered signal $d(t,s_n)$ that is indexed by fast-time (range) $t$ and slow-time (pulse number) $s_n$:
\begin{equation}
    d(t,s_n)=\mathcal F[\rho](t,s_n)+\epsilon_n(t,s_n),\label{eq:bsc}
\end{equation}
where $\mathcal F(\cdot)$ is a Fourier Integral Operator (FIO) \cite{duistermaat1996fourier, wang2014bistatic, krishnan2011synthetic, yarman2007bistatic, intes2004diffuse} representing the SAR forward model and $\epsilon_n(t,s_n)$ represents additive noise at pulse $n$. We define the FIO for SAR under the Born approximation as follows
\begin{equation}
    \mathcal F[\rho](t,s_n):=\int e^{-j\omega\phi(t,s_n,\xx)}A(\omega,s_n,\bx)\rho(\bx)d\omega d\bx,\label{eq:dtsn}
\end{equation}
where $\omega$ is the fast-time frequency, $\phi$ is a phase function corresponding to travel time, and $A$ is an amplitude function that varies slowly with respect to $\omega$ \cite{mason2017deep}. Without loss of generality, we assume $A\equiv1$. For monostatic SAR, we define the phase function as:
\begin{align}
    \phi(t,s_n,\xx)=t-\frac{2}{c}\lVert\gamma(s_n)-\xx\rVert,\label{eqn:phase}
\end{align}
where $c$ represents the speed of light, $\gamma(s_n)\in\R^3$ denotes the platform's 3D position at pulse $n$, and $\lVert\cdot\rVert$ is the Euclidean norm. 


\section{Edge Mapping}
\label{sec:edge-mapping}
Without loss of generality, we set $A(\omega,s_n,\bx)\equiv1$ and $\Psi(\bx)\equiv0$. Assume the scene diameter is much smaller than the range of the antenna and that the origin of the coordinate system is at the center of the scene. Then, in the far-field, we can express \eqref{eqn:phase} as
\begin{align}
    \phi(t,s_n,\xx)\approx t-\frac{2}{c}\left(\lVert\gamma(s_n)\rVert-\bx\cdot\widehat\gamma_\Psi(s_n)\right),
\end{align}
where 
\begin{align}
\widehat\gamma_\Psi(s_n)=\begin{bmatrix}\frac{\gamma_1(s_n)}{\lVert\gamma(s_n)\rVert}&\frac{\gamma_2(s_n)}{\lVert\gamma(s_n)\rVert}&0\end{bmatrix}.
\end{align}
Rearranging the terms in \eqref{eq:dtsn}, we obtain
\begin{align}
    \tilde d(\omega,s_n)\approx\int e^{-j\frac{2\omega}{c}\bx\cdot\widehat\gamma_\Psi(s_n)}\rho(x)dx,\label{eqn:d-tilde}
\end{align}
where
\begin{align}
    \tilde d(\omega, s_n)=\text{FT}\left\lbrace d(\cdot,s_n)e^{j\frac{2\omega}{c}\lVert\gamma(s_n)\rVert}\right\rbrace.\label{eqn:dft-d}
\end{align}
\eqref{eqn:d-tilde} shows that $\tilde d(\omega,s_n)$ is the Fourier transform of $\rho(x)$ evaluated at $\frac{2\omega}{c}\widehat\gamma_\Psi(s_n)$. Let
\begin{align}
\xi=\frac{2\omega}{c}\widehat\gamma_\Psi(s_n).
\end{align} 
Consider the following Laplacian operator: $\lVert\xi\rVert^p,\;p\geq 1$. Then, $\tilde d(\xi)\lVert\xi\rVert^p$ is the Fourier transform of the $\mathcal L*\rho$ where $\mathcal L$ represent the Laplacian in spatial coordinates and $*$ is the convolution operator.

Let $\bar d(\xi_n)=\tilde d(\xi_n)\lVert\xi_n\rVert^p$ and $\xi=\frac{2\omega}{c}\widehat\gamma_\Psi(s_n)$, then
\begin{align}
    \rho_E=\mathcal L*\rho
\end{align}
where $\rho_E$ is the edge-enhanced scene reflectivity function and
\begin{align}
    \bar d(\xi_n)\approx\int e^{-jx\cdot\xi_n}\rho_E(x)dx.\label{eqn:hat-d-xi}
\end{align}
Discretizing $\omega$ into $N_r$ points and stacking $\bar d(\omega_m,s_n)$, we form the observation vector $\dd_n\in\C^{N_r}$. Similarly discretizing $\rho_E$ into $N$ grid points $x_i$ and representing the edge-map by $\br_E\in\C^N$, we form the discrete forward model
\begin{align}
\dd_n=\FF_n\br_E+\boldsymbol{\epsilon}_n,\quad\boldsymbol{\epsilon}_n\sim\mathcal N(0,\RR_n),
\end{align}
where $\FF_n$ is the discrete forward map at slow-time $s_n$.

\section{Sparse SAR}
\label{sec:sparse-sar}

\begin{figure}[ht]
    \centering
    \begin{subfigure}{0.24\textwidth}
        \centering
        \includegraphics[width=\textwidth]{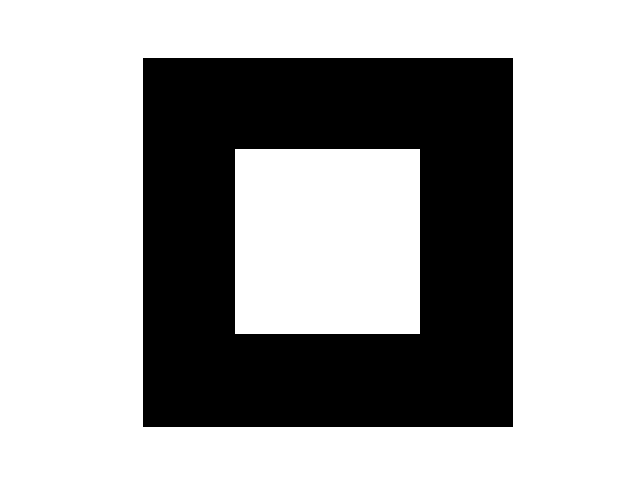}
        \caption{Scene 1 ground truth.}
        \label{fig:1a}
    \end{subfigure}
    \hfill    
    \begin{subfigure}{0.23\textwidth}
        \centering
        \includegraphics[width=\textwidth]{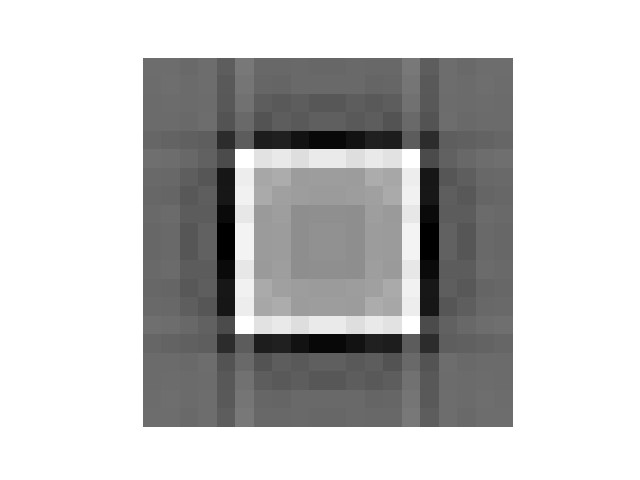}
        \caption{Scene 1 edge mapping.}
        \label{fig:1b}
    \end{subfigure}
    \hfill
    \begin{subfigure}{0.23\textwidth}
        \centering
        \includegraphics[width=\textwidth]{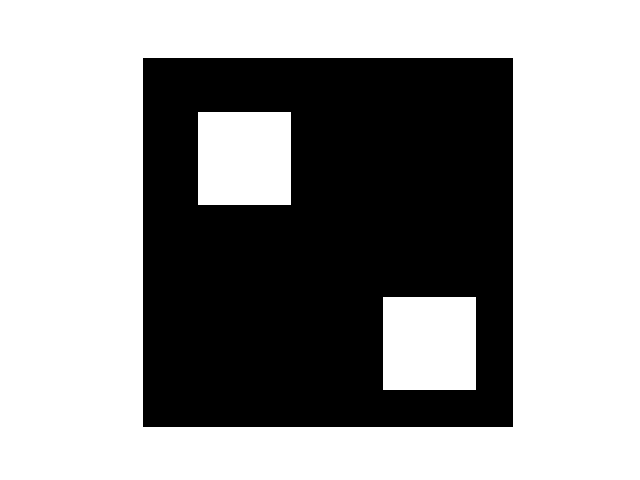}
        \caption{Scene 2 ground truth.}
        \label{fig:1c}
    \end{subfigure}
    \hfill
    \begin{subfigure}{0.23\textwidth}
        \centering
        \includegraphics[width=\textwidth]{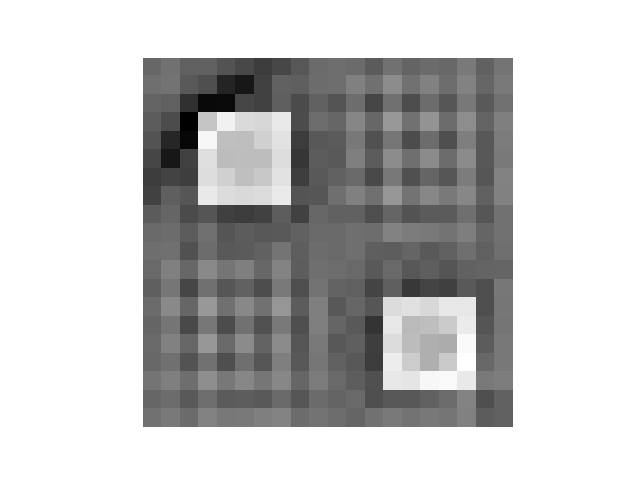}
        \caption{Scene 2 edge mapping.}
        \label{fig:1d}
    \end{subfigure}
    \vspace{10pt}
    \caption{Example scenes with edge-mappings.}
    \label{fig:sample-scenes}
\end{figure}


Since edges in an image are sparse, we represent $\br_E$ by an edgelet dictionary $\HH_E$ and write
\begin{align}
    \br_E\approx\HH_E\cdot\cc
\end{align}
where $\cc$ is sparse. Figure~\ref{fig:sample-scenes} shows an example of two scenes with their derived edge-mappings.

Let $\GG_k=\FF_k\HH$. Through collecting a set of $n$ pulses $\{\dd_k, \GG_k, \RR_k\}_{k=1}^n$, we can fit an optimal $\cc\in\R^M$ to the data through the Weighted LASSO objective \cite{flynn2026online, fista, garrigues2008homotopy} such that
\begin{equation}
    \mathcal J_n(\cc):=\min_\cc\frac{1}{2}\sum_{k=1}^{n}\lVert\dd_k-\GG_k\cc\rVert_{\RR_k^{-1}}^2+\lambda||\cc||_1.\label{eqn:objective-function-j}
\end{equation}
Let $f(\cc)$ be defined as
\begin{equation}
f(\cc)=\frac{1}{2}\sum_{k=1}^n\lVert\dd_k-\GG_k\cc\rVert_{\RR_k^{-1}}^2,
\end{equation}
then we define the gradient of $f(\cc)$ w.r.t. $\cc$, $\nabla_\cc f$, to be
\begin{equation}
    \nabla_\cc f(\cc)=\sum_{k=1}^n\GG_k^\dagger\RR_k^{-1}(\GG_k\cc-\dd_k).
\end{equation}
Furthermore, a valid Lipschitz constant for $\nabla_\cc f$ is
\begin{equation}
    L\geq \left\lVert\sum_{k=1}^n\GG_k^\dagger\RR_k^{-1}\GG_k\right\rVert_2,
\end{equation}
where $||\cdot||_2$ denotes the spectral norm \cite{xiao2009dual, asif2010dynamic}. 

This sparse interpretation, coupled with the sufficient statistics $\AAA_n,\;\bb_n$:
\begin{align}
    \Delta\AAA_n=\GG_n^\dagger\RR_n^{-1}\GG_n,\quad\Delta\bb_n=\GG_n^\dagger\RR_n^{-1}\dd_n,
\end{align}
where $(\cdot)^\dagger$ denotes the Hermitian transpose operator and
\begin{align}
    \AAA_n=\AAA_{n-1}+\Delta\AAA_n,\quad\bb_n=\bb_{n-1}+\Delta\bb_n,
\end{align}
are used in our Online Fast Iterative Shrinkage-Thresholding Algorithm (Online FISTA)\cite{flynn2026online} and a key component for mapping SAR pulses to an interpretable, lower-dimension RL state.

\section{CONCLUSION}
\label{sec:conclusion}
This paper presents a direct, online CS-based edge-mapping approach for SAR imaging. We introduced the mono-static SAR problem, generalized it to the sparse SAR problem using the context of our algorithm, Online FISTA\cite{flynn2026online}, and the discussed the structuring of the edge-mapping on the received signal. Benefits from structuring the algorithm as proposed include better computation and memory efficiency, easier post-processing, and online parameter tuning. Furthermore, by using edge-mapping with CS, we promote a sparser representation reducing the pulse requirement while retaining performance for post-processing techniques. Future work involves more refined dictionary declaration, improvement as to the direction of the edge operator, an increased robustness to noise, and combining the work with downstream tasks such as ATR.


\acknowledgments 
This work was supported by the Air Force Office of Scientific Research
(AFOSR) under the agreement FA9550-23-1-0604.
 
\bibliography{report} 
\bibliographystyle{spiebib} 

\end{document}